\documentclass{article}
\usepackage{amsmath,epsfig}
\usepackage[preprint]{spconfa4}
\usepackage{graphicx}
\usepackage{url}
\usepackage{multirow}
\usepackage{color}
\usepackage{lipsum}

\let\OLDthebibliography\thebibliography
\renewcommand\thebibliography[1]{
  \OLDthebibliography{#1}
  \setlength{\parskip}{0pt}
  \setlength{\itemsep}{0pt plus 0.3ex}
}

\graphicspath{{figs/}}

\begin{document}\sloppy

\def\x{{\mathbf x}}
\def\L{{\cal L}}

\title{SN-Graph: a Minimalist 3D Object Representation for Classification}
\name{Siyu Zhang$^{\dagger}$, Hui Cao$^{\dagger}$, Yuqi Liu$^{\dagger}$, Shen Cai$^{\dagger,\ast}$, Yanting Zhang$^{\dagger}$, Yuanzhan Li$^{\dagger}$ and Xiaoyu Chi$^{\ddagger}$}
\address{$^{\dagger}$School of Computer Science and Technology, Donghua University, Shanghai, CHINA \\ $^{\ddagger}$Qingdao Research Institute of Beihang University, Qingdao, CHINA \\
$^{\ast}$Corresponding author: hammer\_cai@163.com}

\maketitle

\newcommand\blfootnote[1]{%
\begingroup 
\renewcommand\thefootnote{}\footnote{#1}%
\addtocounter{footnote}{-1}%
\endgroup 
}

\begin{abstract}
    Using deep learning techniques to process 3D objects has achieved many successes. However, few methods focus on the representation of 3D objects, which could be more effective for specific tasks than traditional representations, such as point clouds, voxels, and multi-view images. In this paper, we propose a Sphere Node Graph (SN-Graph) to represent 3D objects. Specifically, we extract a certain number of internal spheres (as nodes) from the signed distance field (SDF), and then establish connections (as edges) among the sphere nodes to construct a graph, which is seamlessly suitable for 3D analysis using graph neural network (GNN). Experiments conducted on the ModelNet40 dataset show that when there are fewer nodes in the graph or the tested objects are rotated arbitrarily, the classification accuracy of SN-Graph is significantly higher than the state-of-the-art methods.
\end{abstract}

\blfootnote{The work was supported by  National Natural Science Foundation of China (61703092), the Foundation of Key Laboratory of Artificial Intelligence, Ministry of Education, P.R. China  (AI2020003) and Shanghai Municipal Natural Science Foundation (18ZR1401200).}

\vspace{-5mm}
\section{Introduction}
\vspace{-2mm}
Until now, deep learning on processing 2-dimensional (2D) data has achieved a good performance in many tasks, such as object classification, segmentation and detection.
Part of the reason is that the standard image is an ordered data, and the convolutional neural network (CNN) can effectively discover its hidden information.
However, there is no universally admitted data format in 3-dimensional (3D) area.
The 3D data formats commonly used by researchers are multi-view images, voxels and point clouds, which can already be easily processed by neural networks.

Although multi-view images could be fed into CNN-based networks directly, such as MVCNN~\cite{su15mvcnn} and RotationNet~\cite{rotationnet}, the loss of 3D structure information is inevitable.
For voxel based methods~\cite{3dshapenet}~\cite{voxnet}, due to the explosion of computational cost and memory usage, only low-resolution voxels can be provided. 
Point cloud is the most studied data format in 3D classification, segmentation, and detection tasks.
For example, PointNet~\cite{pointnet} first proposes to use a multi-layer perceptron (MLP) network to process point clouds.
However, PointNet cannot pass messages between points and is unable to extract local structural information.
PointNet++~\cite{pointnet++} uses a hierarchical structure, which groups points within three given ranges to learn local features and down-samples the points in each layer.
DGCNN~\cite{dgcnn} finds neighbors of a point by k-Nearest Neighbor (KNN), and leverages features from its neighbors through an edge convolution operation. All the above methods use points sampled by farthest point sampling (FPS) as input and do not consider the rotation-invariant feature extraction.
Recently, learning-based down-sampling of point clouds and pointwise rotation-invariant network are also studied in~\cite{learnToSample} and~\cite{rotainva} respectively.

\begin{figure*}[thp]
 \centering
 \includegraphics[width=0.70\textwidth]{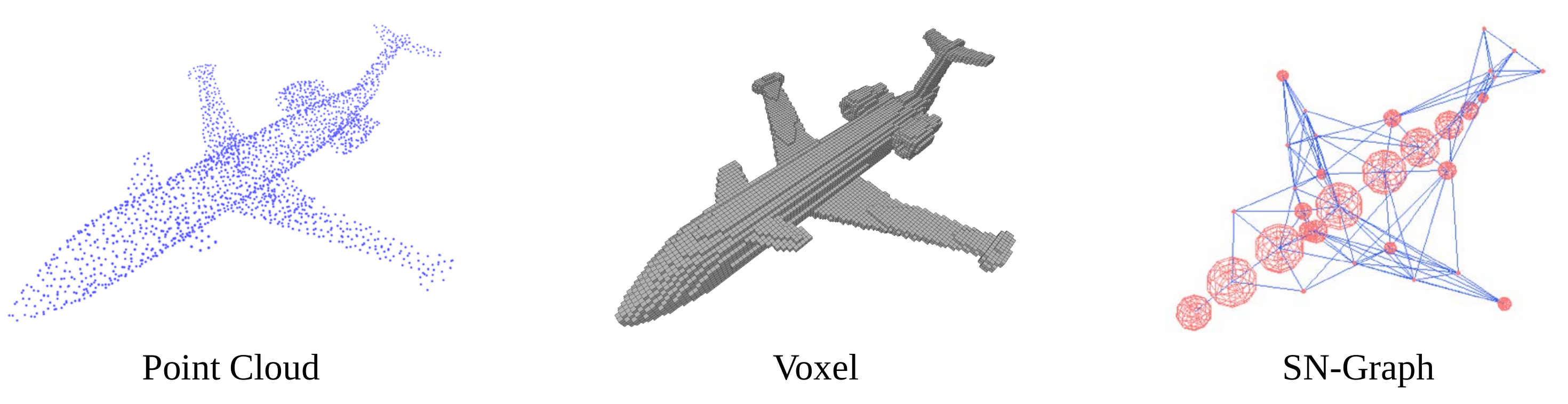}
 \vspace{-5mm}
 \caption{
Point cloud ($1024$), voxel ($128^3$), and the proposed SN-Graph ($32$ nodes) of an airplane are depicted from left to right.}
 \label{fig:1}
 \vspace{-3mm}
\end{figure*}

In this paper, we propose a novel 3D representation named sphere node graph (SN-Graph), which constructs a certain number of internal spheres into a spatial graph.
In more detail, each voxel in voxelized 3D object corresponds to a candidate sphere, whose center is the position of the voxel and the radius is its SDF value. 
A fixed number (e.g.\ 32) of spheres are selected from all internal spheres by a novel sorting criterion considering the distance between two spheres and their radii. 
Then, we check whether the path between two selected spheres exists to determine the edge.
Fig.~\ref{fig:1} shows an example model represented by point cloud, voxel, and our SN-Graph.
Intuitively, SN-Graph contains more spatial information than surface points with the same number of inputs (resolution).
Moreover, building the connections among sphere nodes to form a graph is a bit like extracting the skeleton of an arbitrary 3D object, 
which has demonstrated its capability in human motion analysis~\cite{GraphCNN}.

After constructing the SN-Graph, various well-known graph neural networks (GNNs)~\cite{gcn}~\cite{gat}~\cite{gin}~\cite{deepgcn} could be applied to extract its local and global 3D information. 
We also test the classification performance on ModelNet40 dataset for 3D objects represented by SN-Graph. 
The main contributions of this paper are summarized as follows:
\begin{enumerate}
  \item We propose a minimalist representation for 3D object named SN-Graph, which is geometrically compact and intuitive, even with a few nodes.
  \vspace{-2mm}
  \item We propose a method to automatically construct a graph for any 3D object. The graph can cooperate with multiple GNNs to carry out 3D object classification.
  \vspace{-2mm}
  \item Our method has achieved a classification accuracy that is comparable to the state-of-the-art (SOTA) method~\cite{learnToSample} of learning down-sampling points, and exceeds the previous methods based on FPS points, especially when the resolution is relatively low.
  \vspace{-2mm}
  \item For test dataset under arbitrary rotation, the classification accuracy of SN-Graph is superior than the SOTA method PRIN~\cite{rotainva}, if the proposed rotation-invariant feature of each node is adopted.
\end{enumerate}

\vspace{-5mm}
\section{Related Works}
\vspace{-2mm}
This section will briefly review four directions which are closely related to our research.

\vspace{-3mm}
\subsection{Deep Learning utilizing Signed Distance Field}
\vspace{-2mm}
SDF has been utilized in 3D shape learning for reconstruction or classification task. For example, DeepSDF~\cite{deepsdf} learns an implicit function that maps spatial coordinates to SDF values in 3D reconstruction and completion. BPS~\cite{bps} adopts the SDF values of specially sampled or fixed grid points for classification. InSphereNet~\cite{insphere} firstly uses discrete spheres instead of point clouds for classification. Similar to BPS and InSphereNet, we further explore the classification problem with a simplified graph representation for 3D objects.

\vspace{-3mm}
\subsection{Graph Representation on 3D Point Clouds}
\vspace{-2mm}
There already exist several works exploring to represent 3D objects using graphs~\cite{dgcnn}~\cite{grapgCNN}~\cite{rgcnn}.
For example, the KNN-based graph in DGCNN~\cite{dgcnn} is pretty intuitive and can obtain excellent performance with 2048 points. However, its graph representation is not explicitly defined and the performance drops significantly with the decrease of points number. 
PointGCN~\cite{grapgCNN} simultaneously utilizes KNN and FPS on point clouds to form a graph and apply graph convolution. 
But the graph constructed by using KNN on the FPS points directly still contains some redundant connections.

In sum, to the best of our knowledge, all existing graphs of 3D objects are constructed based on point clouds. 
The neighbour connectivity between points is defined by Euclidean distance (in KNN) or geodesic distance on object surface~\cite{geonet}. 
Thus, their constructed graphs are not as intuitive and geometrically meaningful as the skeleton, which is similar to SN-Graph proposed in this paper. 

\vspace{-3mm}
\subsection{Down-Sampling of Point Clouds}
\vspace{-2mm}
A popular sampling technique is farthest point sampling
(FPS), which is widely used to select a given number of points that are farthest apart from each other~\cite{pointnet}~\cite{pointnet++}. 
There is also a work~\cite{learnToSample} to study how to learn a subset of point cloud to better represent objects for a specific task. 
Although the purpose of reducing the points number (resolution) is similar to our paper, the method of using PointNet and S-Net to predict the point cloud subset is different from our graph representation method. 
Moreover, the learned subset of point clouds has no geometric meaning and will vary from task to task.

\vspace{-3mm}
\subsection{Graph Neural Networks}
\label{sec:GNNsIntro}
\vspace{-2mm}
Graph neural network is a kind of neural network that learns the aggregation relationship between nodes in the graph structure.
From various GNNs, we apply four of them~\cite{gcn}~\cite{gat}~\cite{gin}~\cite{deepgcn} to SN-Graph, respectively taking into account their remarkable characteristics.
GCN~\cite{gcn} introduces convolution operations into graph data that transformed by graph Fourier transform.
Both GAT~\cite{gat} and GIN~\cite{gin} methods use weighted sums to aggregate information of neighbour nodes in spatial domain.
DeepGCN~\cite{deepgcn} further solves the problem of over-smoothing and vanishing gradient due to the large amount of layers in GNNs.



\vspace{-3mm}
\section{Construction of Sphere Node Graph}
\vspace{-2mm}
This section describes the generation process of a SN-Graph, which includes: SDF calculation after voxelizing mesh, sphere nodes selection, and sphere nodes connection. 

\vspace{-3mm}
\subsection{SDF Calculation}
\vspace{-2mm}
Polygon mesh format can be approximately regarded as continuous data, which makes it difficult to calculate the SDF.
For calculation convenience, the mesh could be converted into voxel first, followed by computing the shortest distance from the surface to every internal voxel in the 3D object. We follow the implementation~\cite{edt} to calculate SDF and the voxel resolution is set to $128^3$. Before voxelization, one mesh needs to be normalized according to its longest side. 

\vspace{-3mm}
\subsection{Sphere Nodes Selection}
\label{sec:SNSelection}
\vspace{-2mm}
In order to represent 3D objects by skeleton-like graphs, each node of the graph that also refers to an internal sphere should be selected from voxels.
The first node is selected at the voxel with the largest SDF value, or somewhere closest to the center of the object if there are multiple maximum values. 
This ensures that the first node of the same class of objects is relatively fixed, even if the objects are rotated.
Then, we sort all other voxels to find the next node according to the following distance definition:
\vspace{-2mm}
\begin{equation}
    D(v_i,v_j) =  \underbrace{ ( E(v_i,v_j)-SDF(v_i) ) }_{Global\_Dist}   +   
    \underbrace{ 2*SDF(v_j) }_{Local\_Dist}  
     \label{equ:1} 
     \vspace{-2mm}
\end{equation}
\noindent where $v_i$ and $v_j$ denote two different elements in voxel set $\boldsymbol{V}$; $E(v_i,v_j)$ is the Euclidean distance of the voxels; $SDF(v_i)$ and $SDF(v_j)$ represent the SDF values of these two voxels, respectively.
Assuming that $k$ nodes have been selected, the sorting operation of voxel $v_{k+1}$ corresponding to the next node is expressed by:
\vspace{-2mm}
\begin{equation}
    v_{k+1} = v_j \mid \max \limits_{j\in\boldsymbol{V}} \min \limits_{1\leq i\leq k} D(v_i,v_j) 
   \label{equ:2}
   \vspace{-2mm}
\end{equation}
\noindent By doing so, all nodes can be selected with the designed distance order until a given resolution is achieved.

\begin{figure}[t]
 \centering
 \includegraphics[width=0.45\textwidth]{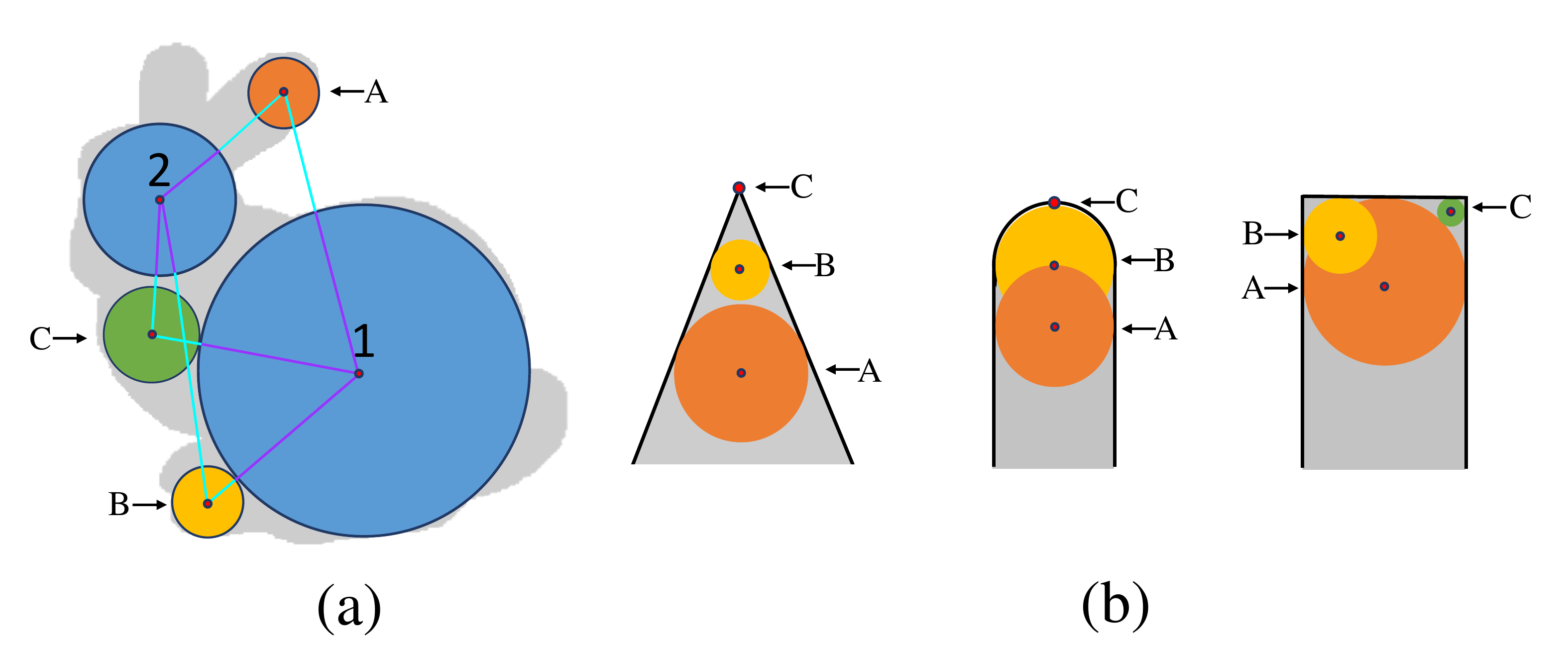}
 \vspace{-4mm}
 \caption{ (a) The influence of the defined global-distance (drawn in cyan) on node selection, which gets rid of the radii of the selected spheres (b) Influence of the defined local-distance on node selection for different local shapes, which enhances the influence of the SDF values of candidate voxels}
 \label{fig:disEqu}
 \vspace{-2mm}
\end{figure}

Eq.~(\ref{equ:1}) actually contains two terms of the distance between the sphere nodes, which we call global-distance ('Global\_Dist') and local-distance ('Local\_Dist').
Fig.~\ref{fig:disEqu} illustrates the influences of these two distances on the next node selection in a planar view. 
In all sub-figures, there are three possible candidate spheres denoted by $A$, $B$ and $C$ of which the centers are drawn in red. Fig.~\ref{fig:disEqu}~(a) also depicts two blue spheres, which denote the first two selected spheres.
The six cyan line segments in Fig.~\ref{fig:disEqu}~(a) denote the global-distance, which is equal to the distance $E(v_i,v_j)$ between two voxels minus the radius (drawn in purple) of a selected sphere. 
At the same time, we restrict the global distance to be greater than zero, which means that one sphere should not contain or intersect another one.
Obviously, according to the global-distance term, the next sphere node will be $A$ (representing a long ear) in Fig.~\ref{fig:disEqu}~(a). 
However, if the distance only depends on $E(v_i,v_j)$, the next sphere will be $B$ (representing a short leg), which is less important than $A$ in terms of the rabbit structure. 
Fig.~\ref{fig:disEqu}~(b) depicts three local regions with different shapes, under the assumption that all selected spheres not drawn here are far away from the local regions.
Under the influence of the coefficient $2$ in local-distance term, the next sphere node to be selected is $C$, $B$ and $A$, from left to right in Fig.~\ref{fig:disEqu}~(b). 
Take the rectangular local shape of the right figure as an example.
Although $A$ is nearest to the selected spheres and has the largest radius, its distance $D(v_i,v_j)$ is $2$ times the radius plus the global-distance term, which is larger than $B$ and $C$.

\begin{figure}[t]
 \centering
 \includegraphics[width=0.45\textwidth]{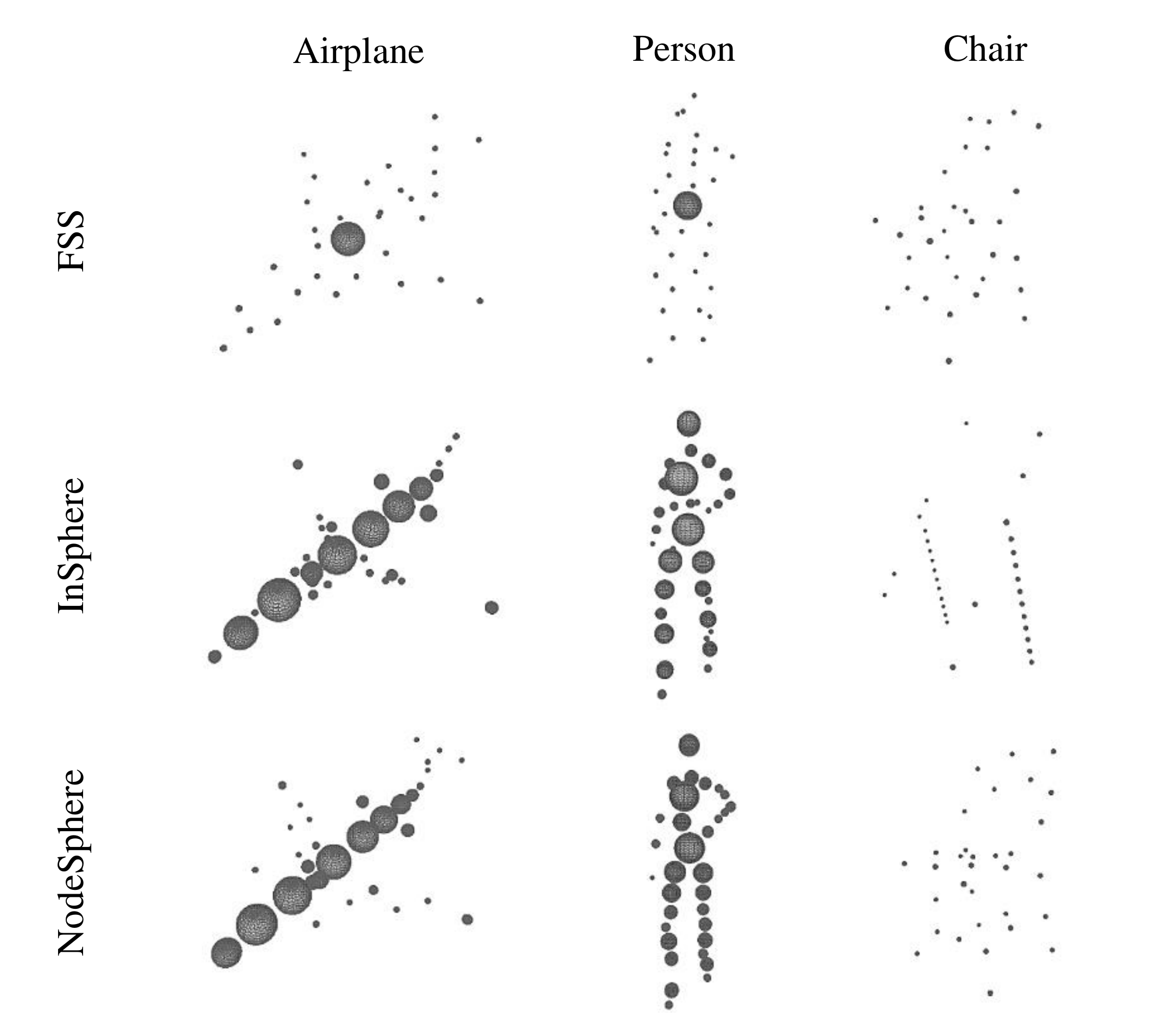}
 \vspace{-5mm}
 \caption{Comparison of different concise representations with the resolution of 32.}
 \label{fig:sampleComp}
 \vspace{-3mm}
\end{figure}

It is worth noting that the proposed spheres selection method (called NodeSphere for simplicity) actually combines FPS and InSphere~\cite{insphere}.
Their differences can be explicitly seen in Fig.~\ref{fig:sampleComp}.
Directly applying FPS to the spheres selection is called farthest sphere sampling (FSS).
FSS does not consider the radius of the sphere node and selects all voxels except the first one on the surface, which are less meaningful than internal spheres with larger radius shown in InSphere and our NodeSphere.
The InSphere sampling process is computationally complex, and tends to select spheres with larger radii (sometimes densely distributed in local area), rather than spheres with slightly smaller radii (evenly distributed everywhere) as in NodeSphere. 
For example, for ‘Airplane’ and ‘Person’ models in Fig.~\ref{fig:sampleComp}, NodeSphere shows more local details, such as the aircraft tail part and the human joints. 
For 'Chair' case, InSphere even loses the entire seat surface.
 
For the NodeSphere representation, we specifically test its performance in the classical methods~\cite{pointnet}~\cite{pointnet++}~\cite{dgcnn}, instead of the original FPS points input. 
In this way, the experimental results shown in Sec.~\ref{sec:exp2} can better reflect the performance difference between the isolated sphere nodes and the connected SN-Graph, which will be discussed in next subsection.

\vspace{-3mm}
\subsection{Sphere Nodes Connection}
\label{sec:SNConnection}
\vspace{-2mm}
After selecting sphere nodes, we need to connect them into a graph.
Inspired by the idea of human joint connection, we propose to construct SN-Graph according to four node connection rules.
\begin{figure*}[thp]
 \centering
 \includegraphics[width=0.9\textwidth]{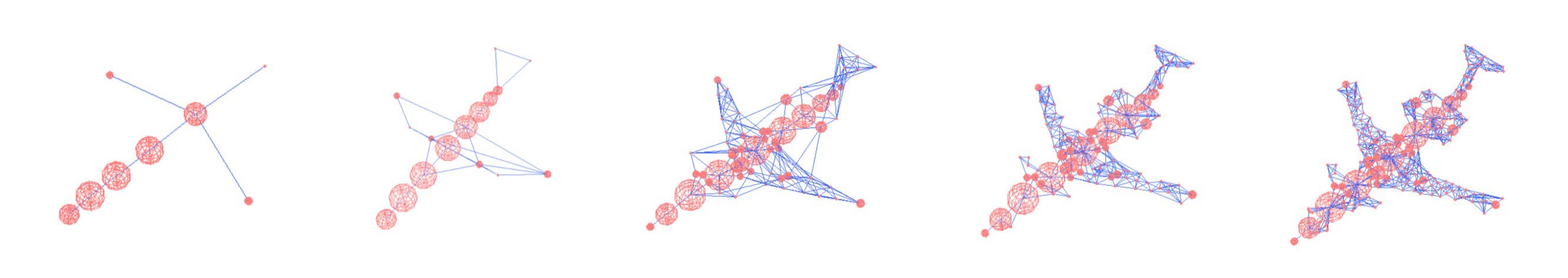}
 \vspace{-8mm}
 \caption{SN-Graphs with different numbers of sphere nodes. From left to right, the resolutions are 8, 16, 64, 128 and 256, respectively. Note that the SN-Graph with 32 nodes has been depicted in Fig.~\ref{fig:1}.}
 \label{fig:different_resolution}
 \vspace{-3mm}
\end{figure*}

\textbf{First, the edge between two nodes should be close enough to the object.} Specifically, $p$ voxels will be sampled uniformly on each edge. When the proportion of voxels outside the object, whose SDF values are less than a threshold $T_d$, is below another threshold $T_p$, the edge will be reserved. In experiments, we set: $p=10$, $T_p=0.7$ and $T_d=0.05$.

\textbf{Second, the edge between two sphere nodes should not intersect another selected sphere.} Obviously, in this manner, two spheres that are far apart will not skip the middle sphere to be connected directly,
even if the edge between them satisfy the first rule.

\textbf{Third, the maximum number of connections from each node to other nodes is limited to $q$.} As the number of connections in dense spheres area may be much larger than it in sparse spheres area, we add this rule to make each node only connect to the nearest $q$ nodes. This operation balances the influence of each node in the graph. Moreover, the simplified graph will be beneficial to local graph convolution. In experiments, we simply set $q=6$ for SN-graphs of all resolutions.

\textbf{Forth, for an isolated sphere node, it needs to be connected to another nearest node.} An isolated sphere node may occur in the distant and curved position of an object. Forcing this node to connect with its nearest node can ensure the integrity of SN-Graph.

The SN-Graphs of an airplane with different resolutions are shown in Fig.~\ref{fig:different_resolution}. 
It can be seen that, even if the graph consists of only 8 nodes, all the SN-Graphs of the airplane are intuitive and distinguishable in shape.
For more implementation details, please see the visualization example of the 32-node airplane in the supplementary material.
In next section, the graph neural networks for object classification using SN-Graph as input will be demonstrated.

\vspace{-3mm}
\section{Graph Neural Networks for SN-Graph}
\vspace{-2mm}
Since the GNN layers we adopt comes from the mature works~\cite{gcn}~\cite{gat}~\cite{gin}~\cite{deepgcn}, this section only illustrates the input feature of sphere node and the network architecture.

\vspace{-3mm}
 \subsection{Input Feature of Sphere Node}
 \label{sec:inputFeature}
 \vspace{-2mm}
The input feature of the sphere node in this paper has two types. The first one is the 3D position and radius of each sphere node, which is abbreviated as PR feature for simplicity. PR feature only adds radius information to commonly used point position input, and will be fed into various GNNs.
Another input feature of each node is rotation-invariant, which includes the angles between any two edges, the distance from this node to any connected node or the origin of the coordinate system, and radii of the node and all connected node. 
Specifically, since a node can connect up to $q=6$ other nodes, cosine values of $6*(6-1)/2=15$ angles can be calculated. 
Therefore, the rotation-invariant feature has $29$ dimensions, which are $15$ cosine values, $7$ distances and $7$ radii. 
If a node has less than 6 connected nodes, the input vector will be filled with 0 to form the 29-dimensional feature.
This feature, abbreviated as ADR feature, will be used in the classification experiment on the rotated test set. The result is shown in Sec.~\ref{sec:rotExp}.
\begin{figure}[t]
\centering
\includegraphics[width=0.48\textwidth]{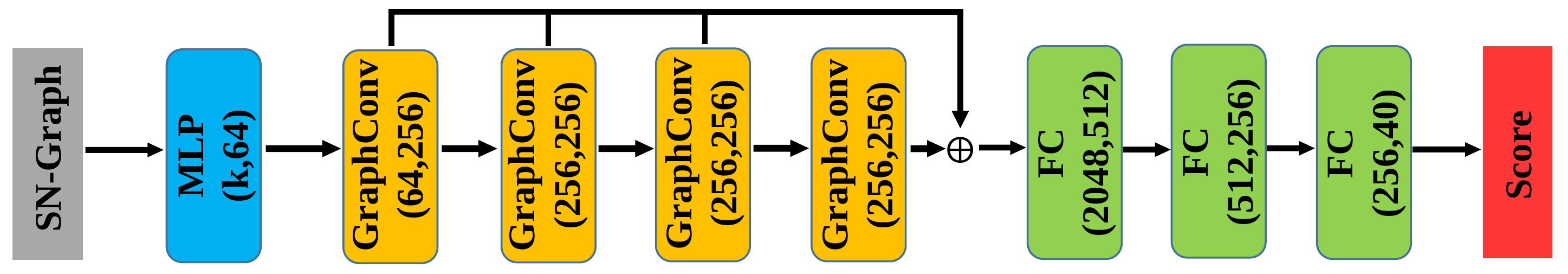}
\vspace{-5mm}
\caption{Architecture of our classification network.}
\label{fig:6}
\vspace{-3mm}
\end{figure}

\vspace{-3mm}
\subsection{Network Architecture}
\vspace{-2mm}
Similar to most graph classification networks, our network follows the design of graph convolution + readout + classifier. 
The architecture of most GNNs we used is shown in Fig.~\ref{fig:6}, except DeepGCN\footnote{As the configurations of DeepGCN~\cite{deepgcn} are different from other GNNs, it cannot be illustrated by Fig.~\ref{fig:6}. Please see the literature and code for details.}. 
We first use an MLP layer or an EdgeConv layer to increase the input feature dimension $k$ to $64$, and then aggregate the features through a 4-layer graph convolution or attention operation.
Then, the concatenation of the global max feature (with dimension of 256) and global mean features (with dimension of 256) of each GraphConv layer is fed to a 3-layer fully connected (FC) network to obtain the classification score.

\vspace{-3mm}
\section{Experiments}
\vspace{-2mm}
The experiments are conducted with the PyTorch Geometric (PyG) library~\cite{pygeometric}, which has implemented many deep learning methods on graph data. Detailed configurations are listed below:
    
   \quad \textit{GPU:} RTX 2080Ti
    
   \quad \textit{Software:} CUDA10.0 + PyTorch1.4 + PyG1.6
    
   \quad \textit{Dataset:} ModelNet40
    
To make a fair comparison, no data augmentation techniques are applied to the training set.
Meanwhile, all networks are trained without fine-tuning.
The resolution of NodeSphere or SN-Graph varies from 8 to 1024.
\textit{All the shown results can be reproduced by our source code in {\color{blue} https://github.com/cscvlab/SN-Graph}}.

\vspace{-2mm}
\subsection{GNNs Performance on SN-Graph}
\vspace{-2mm}
In this experiment, we test the classification performance of four GNNs stated in Sec.~\ref{sec:GNNsIntro} on SN-Graph with resolution $16$, $64$ and $256$.
The results are shown in Table~\ref{tab:gnns_performance}.
All GNNs achieve the similar classification accuracy. 
Among them, GAT on SN-Graph under any resolution shown in the table is a little better than others. This reveals that SN-Graph representation combined with GNNs are effective in the classification of 3D objects.

\vspace{-2mm}
\subsection{Classification Comparison of Various Representations and Networks}
\label{sec:exp2}
\vspace{-2mm}
In this subsection, the classification performance of multiple networks together with the mentioned representations are verified. Five shape representations combined with different networks are tested as follows:
\vspace{-2mm}
\begin{itemize}
\item points sampled by FPS 
\vspace{-1.5mm}

'+' PointNet, PointNet++, DGCNN
\vspace{-2mm}
\item points learned by S-Net~\cite{learnToSample}
\vspace{-1.5mm}

'+' PointNet 
\vspace{-2mm}
\item spheres selected by InSphere~\cite{insphere}
\vspace{-1.5mm}

'+' PointNet
\vspace{-2mm}
\item spheres selected by us (NodeSphere) stated in Sec.~\ref{sec:SNSelection}
\vspace{-1.5mm}

'+' PointNet, PointNet++, DGCNN
\vspace{-2mm}
\item SN-Graph constructed by us stated in Sec.~\ref{sec:SNConnection}
\vspace{-1.5mm}

'+' GAT, DeepGCN
\end{itemize}
\vspace{-2mm}

\begin{figure}[t]
 \centering
 \includegraphics[width=0.48\textwidth]{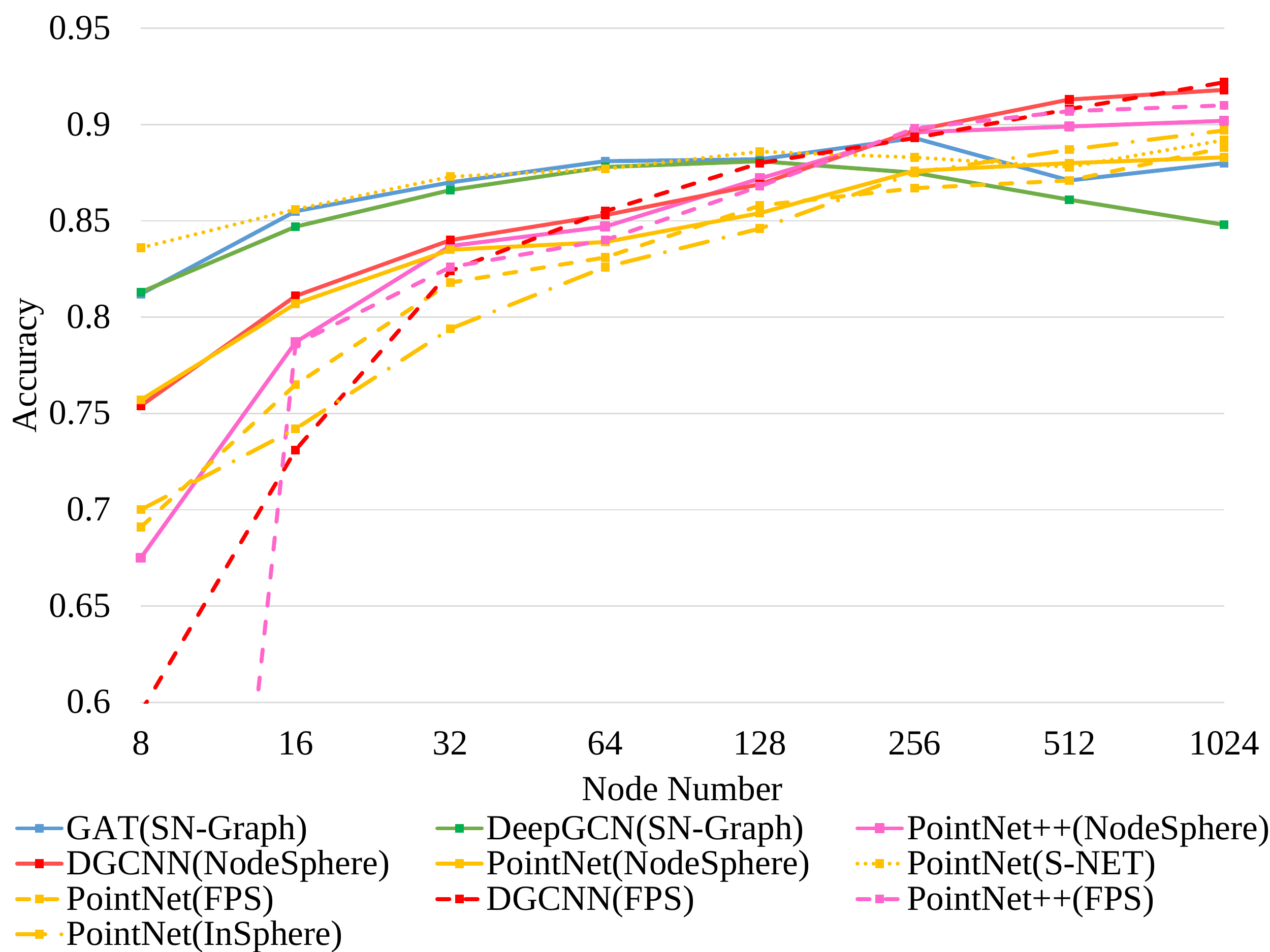}
 \vspace{-3mm}
 \caption{Accuracy comparison of various combinations of representations and networks}
\label{fig:acc}
\vspace{-3mm}
\end{figure}

FPS points are fed into PointNet~\cite{pointnet}, PointNet++~\cite{pointnet++} and DGCNN~\cite{dgcnn}, as the baseline of the classical methods using point cloud representation.
Points learned by S-Net are fed into PointNet that is specially trained for S-Net as the SOTA work~\cite{learnToSample} described.  
Spheres selected by InSphere~\cite{insphere} are also only fed into a PointNet, which is consistent with the original literature.
SN-Graph is combined with two GNNs: GAT and DeepGCN, which we have tested in the above subsection.
To fairly observe the influence of sphere nodes connection, spheres selected by NodeSphere stated in Sec.~\ref{sec:SNSelection} are also fed into PointNet, PointNet++ and DGCNN. 
This setting is equivalent to an ablation experiment of removing graph edges (remaining only nodes) and graph convolution. 
The classification results conducted on ModelNet40 of all the above combination of representations and networks are depicted in Fig.~\ref{fig:acc}. 
Please note that all the results in the figure are trained and inferred by ourselves, except for the results of S-Net which are the direct quote of Table 1 in the literature.
Fig.~\ref{fig:acc} uncovers several meaningful results. 

First, the classification accuracy of two SN-Graph based GNNs drawn by the blue and green solid lines are very close to the SOTA work~\cite{learnToSample} which needs to carefully and separately train its two network modules: S-Net and PointNet.
Meanwhile, the performance of both SN-Graph and S-Net representations drops little with the decrease of graph or point resolution. This fully demonstrates the effectiveness of the proposed SN-Graph representation.
\newcommand{\tabincell}[2]{\begin{tabular}{@{}#1@{}}#2\end{tabular}}
\begin{table}[t]
    \centering
    \begin{tabular}{c|c|c|c|c}
    \multirow{2}{*}{Nodes} &  GAT & GCN  & GIN  & DeepGCN    \\
& ~\cite{gat} & ~\cite{gcn} &  ~\cite{gin} & ~\cite{deepgcn} \\
        \hline
         16 & $\textbf{85.5}$ & $84.5$ & $85.2$ & $84.7$  \\
         64 & $\textbf{88.2}$ & $87.4$ & $87.8$ & $87.8$  \\
        256 & $\textbf{89.2}$ & $87.5$ & $88.7$ & $87.5$  \\
    \end{tabular}
     \vspace{-2mm}
    \caption{GNNs classification accuracy (\%) on SN-Graph}
    \label{tab:gnns_performance}
\end{table}

Second, the accuracy of SN-Graph based methods is obviously higher than the accuracy of NodeSphere or FPS points based methods when the resolution is not larger than 128, no matter what aggregation method is adopted. 
This is explainable as the SN-Graph at low resolution (such as 32) is geometrically appealing and GNNs with four convolution layers are good at handling simple graphs. 
When the resolution becomes higher, the SN-Graph will become messy and lose visual meaning. 
Meanwhile, the ability of 4-layer GNNs to extract features is also getting worse, even weaker than other methods that hierarchically use KNN to aggregate surrounding spheres or points information (see the pink and red lines which denotes the results of PointNet++ and DGCNN, respectively).

Third, the accuracy of FPS points or InSphere is not high under all small resolutions. 
This is as expected because a few FPS points are not only unintuitive for human perception, but also not located at critical positions as pointed out in~\cite{pointnet}~\cite{learnToSample}. 
And the spheres selected by InSphere may be too concentrated to extract local information for a part of the 3D model (see Fig.\ref{fig:sampleComp}). 
Only when the resolution continues to increase (for example, greater than 256), a large number of points or spheres cause information redundancy.
At this time, the dense representation covers the critical points or spheres, and prevents the loss of local information. 
Combined with a good feature aggregating method, the classification accuracy can be significantly improved.

\vspace{-2mm}
\subsection{Classification on Rotated Shapes}
\label{sec:rotExp}
\vspace{-2mm}
We test several combinations of the PR or ADR feature illustrated in Sec.~\ref{sec:inputFeature} and the networks shown in the above subsection to complete classification on rotated shapes of ModelNet40 test set. 
The SOTA rotation-invariant network PPIN~\cite{rotainva} with 1024 points as input, also participates in this comparison. 
The classification results under 'AR' are shown in Table~\ref{tab:rotation}. 
'AR' means that the training set is not rotated, and the test set is arbitrarily rotated. 
All results with PR input features (the positions and radii of spheres) are significantly worse than them with the rotation-invariant features in PRIN or ADR we proposed. Furthermore, by substituting the input feature of SN-Graph (with 256 nodes) from PR to ADR, the GAT method is superior to the SOTA method PRIN. 
\begin{table}[t]
\centering
\begin{tabular}{lcc}
\hline
Network  & Input feature  &  Accuracy (\%)\\
\hline
PointNet  & PR of NodeSphere(64) & $11.3$    \\
PointNet  & PR of NodeSphere(256) & $11.3$   \\
PointNet++  & PR of NodeSphere(64) & $15.2$     \\
PointNet++  & PR of NodeSphere(256) & $16.1$   \\
DeepGCN  & PR of SN-Graph(64) & $21.4$     \\
DeepGCN  & PR of SN-Graph(256) & $23.4$   \\
GAT  & PR of SN-Graph(64) & $19.6$    \\
GAT  & PR of SN-Graph(256) & $21.6$   \\
\hline
PRIN &  xyz of points(1024)  & $70.4$ \\
\hline
DeepGCN  & ADR of SN-Graph(64) & $61.7$     \\
DeepGCN  & ADR of SN-Graph(256) & $63.5$    \\
GAT  & ADR of SN-Graph(64) & $68.2$     \\
GAT  & ADR of SN-Graph(256) & $\textbf{72.7}$    \\
\hline
\end{tabular}
\vspace{-2mm}
\caption{Classification accuracy on ModelNet40 under 'AR'. 'AR' means that the training set is not rotated, and the test set is arbitrarily rotated. GAT with ADR feature as input can be superior to the SOTA method PRIN.}
\label{tab:rotation}
\vspace{-3mm}
\end{table}

\vspace{-3mm}
\section{Conclusion}
\vspace{-2mm}
In this article, we introduce the SN-Graph, a novel 3D object representation, which is geometrically attractive and intuitive to human perception.
Compared with the previous methods based on FPS points, the constructed SN-Graph combined with a variety of graph neural networks performs more effectively in 3D object classification tasks, especially when the number of nodes is not large.
The classification accuracy on the ModelNet40 dataset can also be compared with the accuracy of the SOTA learning-based points representation method~\cite{learnToSample}.
In addition, since the features of SN-Graph, such as angles between two edges, distance between two nodes and radius of one node, are invariant under any rotation, our method obtains a higher classification accuracy than the SOTA technology PRIN~\cite{rotainva} under the rotated test set of ModelNet40.
In the future, it will be a meaningful attempt to apply SN-Graph to other applications, such as 3D object segmentation and scene understanding. Besides, how to improve GNN to process complex graphs (containing a large number of nodes) is also worth exploring.

\small
\bibliographystyle{IEEEbib}
\bibliography{references}
\end{document}